\documentclass[letterpaper, 10 pt, conference]{ieeeconf}  % Comment this line out if you need a4paper

\IEEEoverridecommandlockouts                              % 

\usepackage{hyperref}

\usepackage{graphicx} % for pdf, bitmapped graphics files
\usepackage{amsmath} % assumes amsmath package installed
\usepackage{amssymb}  % assumes amsmath package installed
\usepackage{bm}
\usepackage{algorithm}
\usepackage[noend]{algpseudocode} % algorithm
\usepackage{makecell}

\usepackage{subcaption}
\usepackage{booktabs}

\usepackage{color}

\DeclareMathOperator*{\argmin}{arg\,min}

\title{\LARGE \bf
Random Fourier Features based SLAM
}

\author{
\authorblockN{Yermek Kapushev\authorrefmark{1}$^,$\authorrefmark{2}$^,$\authorrefmark{3},
Anastasia Kishkun\authorrefmark{1},
Gonzalo Ferrer\authorrefmark{1},
Evgeny Burnaev\authorrefmark{1}\thanks{Evgeny Burnaev was supported by the Russian Foundation for Basic Research grant 21-51-12005 NNIO\_a}}
\authorblockA{\authorrefmark{1}Skolkovo Institute of Science and Technology, Moscow, Russia\\
\authorrefmark{2}Artificial Intelligence Research Institute, Moscow, Russia\\
\authorrefmark{3}Sber AI Lab, Moscow, Russia\\
Email: ekapushev@sberbank.ru, \{a.kishkun, g.ferrer, e.burnaev\}@skoltech.ru}
}

\begin{document}

\maketitle
\thispagestyle{empty}
\pagestyle{empty}

%%%%%%%%%%%%%%%%%%%%%%%%%%%%%%%%%%%%%%%%%%%%%%%%%%%%%%%%%%%%%%%%%%%%%%%%%%%%%%%%
\begin{abstract}
% This  work is dedicated to simultaneous continuous-time trajectory  estimation and  mapping    based on \textit{Gaussian  Processes (GP)}. GP allows representing a  continuous  trajectory efficiently processing sparse  measurements but leaving the possibility to know the robot  pose at  any  given time. Usually, GP for\textit{ Simultaneous Localization and Mapping (SLAM)} has several constraints to receive all the above advantages that also limit the number of possible functions that GP can model. In  this  paper  we  provide  the algorithm based on GP with \textit{Random Fourier Features(RFF)} approximation for SLAM without any constraints. The advantages of RFF for continuous-time SLAM are that we don't rely on Markovian structure of the kernel function, but allow to consider any cross-covariance and at the same time greatly reduce computational complexity by operating in the Fourier space of features.
% The additional speedup can be obtained by limiting the number of features. Finally,  we proposed approach on a synthetic  and  a real benchmark  dataset (Victoria Park).
This work is dedicated to simultaneous continuous-time trajectory estimation and mapping based on \textit{Gaussian  Processes (GP)}. State-of-the-art GP-based models for \textit{ Simultaneous Localization and Mapping (SLAM)} are computationally efficient but can only be used with a restricted class of kernel functions. This paper provides the algorithm based on GP with \textit{Random Fourier Features (RFF)} approximation for SLAM without any constraints. The advantages of RFF for continuous-time SLAM are that we can consider a broader class of kernels and, at the same time, maintain computational complexity at reasonably low level by operating in the Fourier space of features.
The accuracy-speed trade-off can be controlled by the number of features. Our experimental results on synthetic and real-world benchmarks demonstrate the cases in which our approach provides better results compared to the current state-of-the-art. 
\end{abstract}

%%%%%%%%%%%%%%%%%%%%%%%%%%%%%%%%%%%%%%%%%%%%%%%%%%%%%%%%%%%%%%%%%%%%%%%%%%%%%%%%
\section{INTRODUCTION}
Since the last century, probabilistic state estimation has been a core topic in mobile robotics, often as part of the problem of simultaneous localization and mapping \cite{bailey2006simultaneous,durrant2006simultaneous}. Recovery of a robot's position and a map of its environment from sensor data is a complicated problem due to both map and trajectory are unknown as well as the correspondences between observations and landmarks \cite{probabilistic}.
 
 The field of discrete time trajectory estimation and mapping methods is well developed \cite{SAM, 1sam, 2sam, 3sam, 4sam, isam, FASTisam, bayes}.
 However, discrete-time representations are constrained because they are not easily adapted to irregularly distributed poses or asynchronous measurements over trajectories.
%  Such limitations would be well addressed by a continuous-time version of the SAM problem where measurements regulate the trajectory at any time step.
In the time-continuous problem statement, the robot trajectory is a function $\bm{x}(t)$ which corresponds to a robot state at every time $t$. Simultaneous trajectory estimation and mapping (STEAM) presents the problem of estimating this function along with landmark positions \cite{barfoot,barfoot2017}. 
In the work \cite{furgal} they formally derive a 
continuous-time SLAM problem and demonstrate the use of a parametric solution for atypical SLAM 
calibration problems. The use of cubic splines to parameterize the robot trajectory can also be seen in the estimation schemes in~\cite{br, fleps, droeschel2018efficient}.
In the work \cite{tong2013gaussian} the  parametric 
state  representation  was proposed due  to  
practicality and effectiveness.
The advantages of this method are that they can 
precisely model  and  interpolate  asynchronous
data to  recover  a  trajectory  and  estimate 
landmark positions. 
The disadvantages of that algorithm are that it 
requires batch updates and considerable  
computational  problems.
%that are natural for regression.
In  the  work \cite{best}, the  critical  update  to increase the efficiency of  existing  GP approach to solve the  STEAM  problem was introduced.  It 
combines benefits of graph-based SLAM \cite{2sam} and GP-based solution \cite{barfoot2014batch}
to provide a computationally efficient solution to the  STEAM  
problem  even for large datasets.
However,
% in this work GP have several constraints to be able to deal with sparse measurements and provide robot position at any point of interest. 
% The main drawback of the paper is that the proposed GP 
% prior impose constraints on the kernel function and 
% thus limits the number of possible functions that GP can model.
the computational efficiency comes at the cost of constrained class of kernel functions.
They use state-space formulation of GP model.
Fast inference in this case is possible if we impose
Markovian structure on the trajectory: it is supposed that two points on the trajectory are conditionally 
independent given all other points if these two points are not neighboring.
However, in some cases the accuracy of the trajectory
estimate can be increased by adjusting the estimate
at the current point using all previous points in the trajectory, especially when the observations
contain a considerable amount of noise.

In recent years a lot of effort has been put to develop large-scale GP models without any constraints on the kernel function~\cite{rudi2017falkon, wang2019exact, munkhoeva2018quadrature}.
There are two main approaches to scale up the GP model.
The first one is based on Nystr{\"o}m approximation \cite{quinonero2005unifying}.
The idea is to approximate the kernel function using a finite set of basis functions that are based on eigenvectors of the kernel matrix.
This approach is data-dependent and needs updating the basis function when new observations arrive.
Another set of methods is based on Random Fourier Features (RFF) \cite{rahimi2008random}.
In these approaches the basis functions depend solely on the kernel function and independent of the data set.
It provides additional computational benefits and is more attractive for SLAM problems.

The {\em contributions} of this paper are as follows.
We develop random features-based SLAM approach.
It uses a low-rank approximation of the kernel matrix which is dense and, therefore,
does not assume the conditional independence of the points on the trajectory.
We show that in certain situations removing the independence constraint allows to improve the quality of trajectory estimation.
Also, low-rank structure of the approximation
allows to maintain the computational complexity at
reasonably low level.

The paper is organized as follows.
Section \ref{sec:gp} provides background on Gaussian Processes and random features-based approximation.
In Section \ref{sec:rff_slam} we describe the proposed RFF-based SLAM.
Section \ref{sec:experiments} contains experimental results demonstrating how the method works in practice in comparison with the competing approach.
Finally, Section \ref{sec:conclusion} concludes the paper.
% In this paper we develop random features based SLAM
% approach.
% We build a low-rank approximation of the kernel matrix. It is dense, so we do not assume the conditional independence of the points on the trajectory.
% At the same time we maintain reasonable computational complexity by limiting the number of RFF features.
% We conduct several experiments and discuss the results of running the method on synthetic data and well-known real-world benchmarks. 

\section{GAUSSIAN PROCESSES}
\label{sec:gp}
One of the most efficient tools for approximating smooth functions is the Gaussian Process (GP)
Regression \cite{rasmussen2004gaussian,burnaev2016regression}.
GP regression is a Bayesian approach where a prior distribution over functions is assumed to be a Gaussian Process, i.e. $y =  f(\bm{x}) + \varepsilon$ with $f\sim\mathcal{GP}(\mu(\bm{x}),k(\bm{x},\bm{x}'))$ and white noise $\varepsilon\sim\mathcal{N}(0,\sigma^2_{noise})$, so that
\[
  \mathbf{y} \, | \, \mathbf{X} \sim \mathcal{N}(\boldsymbol{\mu}, \, \mathbf{K}_f + \sigma_{noise}^2\mathbf{I}),
\]
where $\mathbf{y} = (y_1, y_2, \ldots, y_N)$ is a vector of outputs,
$\bm{X} = (\bm{x}_1^{\top}, \bm{x}_2^{\top}, \ldots, \bm{x}_N^{\top})^{\top}$ is a matrix of inputs,
$\bm{x}_i \in \mathbb{R}^d$,
$\sigma_{noise}^2$ is a noise variance,
$\boldsymbol{\mu} = (\mu(\bm{x}_1), \mu(\bm{x}_2), \ldots, \mu(\bm{x}_N))$ is a mean vector modeled by some function $\mu(\bm{x})$,
$\mathbf{K}_f = \{ k(\bm{x}_i, \bm{x}_j) \}_{i, j = 1}^N$ is a covariance matrix for some a priori selected covariance function $k$ and
$\mathbf{I}$ is an identity matrix.
An example of such a function is a Radial Basis Function (RBF) kernel
\[
k(\bm{x}, \bm{x}') = \exp \left (
- \frac{1}{2} \sum_{i=1}^d \left (
\frac{\bm{x}^{(i)} - \bm{x}'{}^{(i)}}{\sigma_i}
\right )^2
\right ),
\]
where $\sigma_i, i = 1, \ldots, d$ are parameters of the kernel
(hyperparameters of the GP model) and
$\bm{x}^{(i)}$ is an $i$-th element of vector $\bm{x}$.
The hyperparameters should be chosen according to the given data set.

For a new unseen data point $\bm{x}_*$ the conditional distribution of $f(\bm{x})$ given $(\bm{y},\bm{X})$ is equal to
\begin{equation}
\label{eq:gp_posterior}
\begin{gathered}
    \hat{f}(\bm{x}_*)  \sim \mathcal{N}\left (\hat{\mu}(\bm{x}_*), \hat{\sigma}^2(\bm{x}_*) \right ),
    \\
    \hat{\mu}(\bm{x}_*) = \mu(\bm{x}_*) + \mathbf{k}(\bm{x}_*)^\top \mathbf{K}^{-1}\left (\mathbf{y} - \bm{\mu} \right ),
    \\
    \hat{\sigma}^2(\bm{x}_*) = k(\bm{x}_*, \bm{x}_*) - \mathbf{k}(\bm{x}_*)^\top \mathbf{K}^{-1} \mathbf{k}(\bm{x}_*),
\end{gathered}
\end{equation}
where $\mathbf{k}(\bm{x}_*) = (k(\bm{x}_*, \bm{x}_1), \ldots, k(\bm{x}_*, \bm{x}_N))^T$ and
$\mathbf{K} = \mathbf{K}_f + \sigma_{noise}^2\mathbf{I}$.

The runtime complexity of the construction of the GP regression model is $\mathcal{O}(N^3)$ as we need to calculate the inverse of $\mathbf{K}$.

\subsection{Random Fourier Features}
To approach the computational complexity of building a GP model we use {\em Random Fourier Features} (RFF).
The idea behind RFF is in Bochner's theorem \cite{rudin2017fourier} stating that any shift-invariant kernel $k(\bm{x}, \bm{y}) = k'(\bm{x - y})$ is a Fourier transform of a non-negative measure $p(\bm{w})$, i.e. 
\[
    k'(\bm{x - y}) = \int p(\bm{w}) e^{j\bm{w^\top(x - y)}}d\bm{w}.
\]
Then the integral and, therefore, the kernel can be approximated as follows
\begin{equation}
\label{eq:bochner_theorem}
    k'(\bm{x - y}) \approx \phi(\bm{x})^\top \phi(\bm{y}),
\end{equation}
where
\begin{equation}
\label{eq:mapping}
    \phi(\bm{x}) = \sqrt{\frac{2}{D}}\begin{bmatrix}
        \cos (\bm{w}_1^\top \bm{x}) \\
        \sin (\bm{w}_1^\top \bm{x}) \\
        \cdots \\
        \cos (\bm{w}_{D / 2}^\top \bm{x}) \\
        \sin (\bm{w}_{D / 2}^\top \bm{x}) \\        
    \end{bmatrix}, 
    \quad \bm{w}_j \sim p(\bm{w}).
\end{equation}
In this case the kernel matrix has a low-rank representation,
$
    \mathbf{K}_f \approx \Psi \Psi^{\top}
$,
$
    \Psi = \|\phi(\bm{x}_i)^\top\|_{i = 1}^N \in \mathbb{R}^{N \times D}.
$
Therefore, $\mathbf{K}^{-1}\bm{x}$
can be efficiently calculated in $\mathcal{O}(ND^2)$
using Sherman-Morrison-Woodbury matrix identity, i.e.
linearly in the number of observations.
Constant $D$ --- the number of features --- is a hyperparameter of the algorithm, that controls the accuracy of the kernel matrix approximation.
Alternatively, after we find finite-dimensional 
feature map that is used to approximate the kernel
function as in \eqref{eq:bochner_theorem},
we can work in weight space view using
$\phi(\bm{x})$ (see~\cite{rasmussen2004gaussian}).
There are several approaches (e.g., \cite{choromanski2017unreasonable, munkhoeva2018quadrature}) that both improves the quality of RFF and reduces the complexity of generating RFF to $\mathcal{O}(d \log d)$, where $d$ is the dimensionality of $\bm{x}$.
Such computational complexity makes RFF a good candidate to be used for SLAM.

\section{SLAM}
\label{sec:slam}
In this paper we consider the following SLAM problem.
Let $\bm{l}$ be a map consisting of $L$ landmarks.
Let $\mathbf{z} = \begin{bmatrix}z(t_1) & \cdots & z(t_N)\end{bmatrix}$ and
$\mathbf{u} = \begin{bmatrix}u(t_1) & \cdots & u(t_N)\end{bmatrix}$
be vectors of measurements and controls at time steps $t_1, \ldots, t_N$.
Our goal is to estimate the posterior probability of the robot poses
$\mathbf{x} = \begin{bmatrix} \bm(x)(t_0) & \cdots & \bm(x)(t_N)\end{bmatrix}$ and landmarks given the measurements and controls:
\begin{equation}\label{full}
    \quad p\left(\mathbf{x}, \bm{l} | \mathbf{z}, \mathbf{u}\right).
\end{equation}

% is not restricted to this setting and can be applied in different variations of SLAM.
% Online SLAM (\ref{online}) involves estimating the posterior over the current pose along the map:
% \begin{equation} \label{online}
% p\left(\bm{x}_{t}, \bm{l} | \mathbf{z}, \mathbf{u}\right),
% \end{equation}
% where $\bm{x}_{t}$ is the pose at time $t, \bm{l}$ is the map (in this work, we consider $\bm{l}$ to be the map of landmarks), and $\mathbf{z} = \begin{bmatrix}\bm{z}_(t_1) & \cdots \bm{z}(t_N)\end{bmatrix}$ and $\mathbf{u} = \begin{bmatrix}\bm{u}_(t_1) & \cdots \bm{u}(t_N)\end{bmatrix}$ are the measurements and controls, respectively. 
% In full SLAM (\ref{full}), we seek to calculate a posterior over the entire path $\mathbf{x} = \begin{bmatrix}\bm{x}_(t_0) & \cdots \bm{x}(t_N)\end{bmatrix}$ along with the map, instead of just the current pose $\bm{x}_{t}$:
% \begin{equation}\label{full}
%     \quad p\left(\mathbf{x}, \bm{l} | \mathbf{z}, \mathbf{u}\right).
% \end{equation}

\subsection{RFF-SLAM}
\label{sec:rff_slam}

We use GP regression with RFF to estimate the state variables corresponding to trajectory and map (landmarks).
Our model is as follows
\begin{equation}
\label{eq:gp_slam_model}
    \begin{aligned}
        \bm{x}(t) &\sim \mathcal{GP}(\bm{\mu}_x(t), \bm{k}(t, t')), \\
        \bm{l} &\sim \mathcal{N}(\bm{\mu}_l, \mathbf{L}), \\
        \bm{z}_i &= \bm{h}\left (
        \begin{bmatrix}
            {\bm x}(t_i) \\
            {\bm l}
        \end{bmatrix}
        \right ) + \bm{n}_i,
    \end{aligned}
\end{equation}
where ${\bm x}(t)$ is a state of the robot at timestamp $t$,
${\bm l}$ is a vector of $M$ landmarks,
$\bm{h}(\cdot)$ is a non-linear measurement model,
$\bm{n}_i \sim \mathcal{N}(\bm{0}, \mathbf{R}_i)$ is measurement noise,
$t_1, \ldots, t_N$ is a sequence of measurement times and
$(\bm{\mu}_l, \mathbf{L})$ are prior mean and the covariance of the landmarks positions.

The paper \cite{tong2013gaussian} uses GP for SLAM and provides the main equations to solve the problem.
We follow their approach with the difference that we utilize RFF approximation of the RBF kernel.
For the RBF kernel its Fourier transform is defined by $p(\bm{w})$ being a Gaussian distribution $\mathcal{N}\left (0, \frac{1}{\sigma_l^2}\mathbf{I} \right )$.
Explicit mapping \eqref{eq:mapping} allows working in
weight-space view
\[
{\bm x}(t) = {\bm \mu}_x(t) + \begin{bmatrix}
        \phi_1(t)^\top {\bm b}_x^{(1)} \\
        \cdots \\
        \phi_d(t)^\top {\bm b}_x^{(d)} \\
    \end{bmatrix}
    + \varepsilon,
\]
where ${\bm b}_x^{(m)} \sim \mathcal{N}({\bm \mu}_b^{(m)}, \mathbf{K}_m), m = 1, \ldots, d$,
$d$ is the state size,
${\bm b}_x^{(m)} \in \mathbb{R}^D$,
$\phi_m({\bm x})$ is a feature map for the $m$-th state variable,
$\mathbf{K}_m \in \mathbb{R}^{D\times D}$ is the prior covariance matrix of the random variable
${\bm b}_x^{(m)}$ and
$\varepsilon \sim \mathcal{N}(0, \sigma^2 \mathbf{I})$ is a Gaussian noise with variance $\sigma^2$.
In principle, the same feature map can be used for all variables,
however, it can be reasonable to use different features
(corresponding to different kernels) to model different types of variables
(for example, coordinates on the map and angles).
% The distribution of weights ${\bm b}_x$ is Gaussian, ${\bm b}_x \sim \mathcal{N}(\bm{\mu}_{b_x}, \mathbf{K}_b)$,
% $\mathbf{K}_b = \begin{bmatrix}
%     \mathbf{K_1} & & \\
%     & \ddots & \\
%     & & \mathbf{K}_d \\
% \end{bmatrix}$,
% $\mathbf{K}_i \in \mathbb{R}^{D \times D}$.

Let us denote 
\begin{equation}
\label{eq:system_variables}
\begin{gathered}
    {\bm b} = \begin{bmatrix}
        {\bm b}_x^{(1)} &
        \cdots &
        {\bm b}_x^{(d)} &
        {\bm l}
    \end{bmatrix}^\top,
    \\
    {\bm \mu} = \begin{bmatrix}
        {\bm \mu}_b^{(1)} &
        \cdots &
        {\bm \mu}_b^{(d)} &
        {\bm \mu}_l
    \end{bmatrix}^\top,
    \\
    \mathbf{K} = {\rm diag}\left (
        \mathbf{K}_1, 
        \ldots,
        \mathbf{K}_d
    \right ),
    \mathbf{P} = {\rm diag}\left (
        \mathbf{K}, \mathbf{L}
    \right ),
    \\
    {\bm \Phi}_i = {\rm diag} \left ( 
        \phi_1(t_i)^\top,
        \ldots,
        \phi_d(t_i)^\top,
        \mathbf{I}_{2M}
    \right )
    \\
    \mathbf{R} = {\rm diag} \left (
        \mathbf{R}_1,
        \ldots,
        \mathbf{R}_N
    \right ).
\end{gathered}
\end{equation}
Now to obtain both the robot states and landmarks position
${\bm b}$ we employ maximum a posteriori (MAP) estimate
\begin{align}
\label{eq:map_equation}
    p({\bm b} | {\bm z})
    %&=
    %\frac{p({\bm z} | {\bm b}) p({\bm b})}{p({\bm z})} \nonumber
    %\\
    &\propto
    -\frac12 \left (
        \sum_{i=1}^N
        \|{\bm z_i} - {\bm h}({\bm \Phi}_i {\bm b})\|_{\mathbf{R}_i}^2
        + \|{\bm b} - {\bm \mu}\|_{\mathbf{P}}^2
    \right ) \nonumber \\
    & \rightarrow \max_{{\bm b}}.
\end{align}
To solve the problem we do the following.
Suppose, that we have an initial guess $\bar{\bm{b}}$.
We update the estimate iteratively by finding the optimal perturbation vector $\delta \bm{b}^*$ for the linearized measurement model.
Namely, we apply the first order Taylor expansion to the measurements model
\[
    \bm{h}({\bm \Phi}_i {\bm b}) \approx \bm{h}({\bm \Phi}_i \bar{\bm{b}}) + \mathbf{H}_i \delta \bm{b}, \quad 
    \mathbf{H}_i = \left . \frac{\partial \bm{h}({\bm y})}{\partial \bm{y}} \right |_{{\bm y} = {\bm \Phi}_i \bar{\bm{b}}}.
\]
Plugging linearized measurement into \eqref{eq:map_equation}
we obtain the following optimization problem
\begin{align*}
    \delta \bm{b}^* = \argmin_{\delta \bm{b}}
    \frac12 \left ( \sum_{i = 1}^N \right . &
    \|\bm{z}_i - \bm{h}({\bm \Phi}\bar{\bm{b}}) -
    \mathbf{H}_i{\bm \Phi}_i\delta \bm{b}\|_{\mathbf{R}_i}^2 
    \\
    &\left. + \|\bar{\bm{b}} + \delta \bm{b} - \bm{\mu}\|_{\mathbf{P}}^2
    \vphantom{\sum_{i=1}^N}
    \right ).
\end{align*}
The solution is given by
\begin{equation}
\label{eq:state_update}
\begin{gathered}
    \delta {\bm b}^* = \mathbf{A}^{-1}{\bm g},
    \\
    \mathbf{A} = \sum\limits_{i=1}^N
        {\bm \Phi}_i^\top \mathbf{H}_i^\top \mathbf{R}_i^{-1} \mathbf{H}_i {\bm \Phi}_i
        + \mathbf{P}^{-1},
    \\
    {\bm g} = \sum\limits_{i=1}^N
        {\bm \Phi}_i^\top \mathbf{H}_i^\top \mathbf{R}_i^{-1} \left (
    {\bm z}_i - {\bm h}(\Phi_i \bar{{\bm b}})
    \right ) + \mathbf{P}^{-1} \left (
    \bar{{\bm b}} - {\bm \mu}_b
    \right ).
\end{gathered}
\end{equation}

We update the model parameters
$\bar{{\bm b}} \gets \bar{{\bm b}} + \delta {\bm b}^*$,
then update all the matrices and vectors in \eqref{eq:system_variables}
and repeat the procedure predefined number of iterations or until convergence.
The described approach is known as Gauss-Newton method for
non-linear least squares problems, and it is used in
\cite{tong2013gaussian, barfoot}.
It does not guarantee convergence, so in this work
we apply Levenberg-Marquardt approach.
It modifies the system
\begin{equation}
\label{eq:levenberg_marquardt}
    \delta {\bm b}^* = \left (
        \mathbf{A} + \lambda {\rm diag} \left ( \mathbf{A} \right )
    \right )^{-1} {\bm g}
\end{equation}
where $\lambda$ is a dampening parameter.
The overall update procedure is summarized in Algorithm~\ref{alg:state_update}.

The size of the system matrix $\mathbf{A}$ is
$(Dd + 2M) \times (Dd + 2M)$ for two-dimensional landmarks.
The top-left block of size $Dd \times Dd $ of the matrix corresponds
to the weights ${\bm b}$ and is typically dense.
The bottom-right block of size $2M \times 2M$ corresponds
to landmarks and it is usually diagonal
(because we assume that landmarks are independent).
Therefore, the cost of solving \eqref{eq:state_update}
is $\mathcal{O}(D^3d^3 + MDd)$ using Schur complement.
However, we use iterative solver and in practice it converges
much faster.
The cost of construction of the matrices in \eqref{eq:state_update}
is $\mathcal{O}(N(D^2d^2 + M))$.
The total complexity is
$\mathcal{O}(ND^2d^2 + NM + D^3d^3 + MDd)$.

When we use the iterative solver that utilizes only matrix-vector 
products,
we do not need to calculate the matrix
$A$ explicitly.
Instead we multiply each term of the sum in \eqref{eq:state_update}
by a vector and then take the sum.
Taking into account that matrices $\mathbf{R}_i$ are (usually)
diagonal, the part of the Jacobi matrix $\mathbf{H}_i$ that
corresponds to derivatives of w.r.t landmarks is block-diagonal,
the complexity of matrix-vector product for one term in the sum
is $\mathcal{O}((M + D)d)$.
The overall complexity of solving the system is
$\mathcal{O}(N(M + D)d k)$, where $k$ is the number of iterations.

\subsection{State prior}
%In GP regression it is common make zero-mean assumption,
% i.e. ${\bm \mu}(t) = {\bm 0}$.
% However,
Having good prior ${\bm \mu}(t)$ is essential
when modeling trajectory with GP, because usually 
shift-invariant kernel functions are used.
The GP with such kernel is most suited to model
stationary functions.
This does not always apply to trajectories.
Non-stationarity can be accounted by non-zero mean function
${\bm \mu}(t)$.
Here we use one of two prior mean functions.
\begin{enumerate}
    \item Motion model:
    % \[
    %     \bm{\mu}_x(t_i) = \mathbf{F}(t_i) \hat{\bm{x}}(t_{i - 1}) + \mathbf{B}(t_i) \bm{u}(t_i),
    % \]
    $
        \bm{\mu}_x(t_i) = \mathbf{F}(t_i) \hat{\bm{x}}(t_{i - 1}) + \mathbf{B}(t_i) \bm{u}(t_i),
    $
    where $\mathbf{F}(t), \mathbf{B}(t)$ are time-dependent
    system matrices.
    We use this model if we have odometry measurements.
    \item Smoothing splines applied to the estimated 
    trajectory with smoothing parameter $0.98$
    (we used De Boor's formulation, 
    see~\cite{bde2001practical}).
    We also use weights that are inverse proportional to the
    data fit error $\|{\bm z}_i - {\bm h}({\bm \Phi}_i {\bm b})\|_{\mathbf{R}_i}$.
    The motivation behind this prior mean model is the following:
    in case of non-stationarity the GP model can oscillate (or it can have other artifacts).
    Smoothing the trajectory reduces such effects.
\end{enumerate}
With a non-zero prior mean for the trajectory we can set
all mean vectors ${\bm \mu}_b^{(i)}$ to zero,
thus, the GP will only correct the errors of the mean
${\bm \mu}(t)$.
The whole trajectory estimate is updated with every new measurement,
so we also update the prior
$\bm{\mu}_x(t_i)$ for all $i=1, \ldots, N$
for each new observation.
\begin{algorithm}
\caption{Update state at measurement times}
\label{alg:state_update}
    \begin{algorithmic}[1]
        \State Initial values $\bar{\bm{b}}$, measurement times $t_1, \ldots, t_N$, measurements $\mathbf{z}$, tolerance $\varepsilon$, max number of iterations $K$
        \State $n \gets 0$
        \Repeat
            \State Using $\bar{\bm{b}}$ update vectors and matrices in \eqref{eq:system_variables}
            \State Calculate update $\delta \bm{b}^*$ by 
            applying \eqref{eq:levenberg_marquardt} to solve \eqref{eq:state_update}
            \State $\bar{\bm{b}} \gets \bar{\bm{b}} + \delta \bm{b}^*$
            \State $n \gets n + 1$
        \Until relative error is less than $\varepsilon$ \textbf{or} $n = K$
    \end{algorithmic}
\end{algorithm}

\section{EXPERIMENTS}
\label{sec:experiments}
In this section, we evaluate our approach
on several synthetic 2D trajectories
as well as real-world benchmarks.
In all our experiments, we consider the state vector to be a 2D pose, i.e.
$
    \bm{x}(t) = \begin{bmatrix}
    x(t) &
    y(t) &
    \alpha(t)
    \end{bmatrix}^\top.
$
We use the range/bearing observation model given by
\begin{equation}
\label{eq:observation_function}
    \bm{h}\left (\begin{bmatrix}
            \bm{x}(t_i) \\
            \bm{l}_j
        \end{bmatrix}
    \right ) =
    \begin{bmatrix}
        \sqrt{(x_j - x(t_i))^2 + (y_j - y(t_i))^2} \\
        {\rm atan2}(y_j - y(t_i), x_j - x(t_i)) - \alpha(t_i)\\
    \end{bmatrix} ,
\end{equation}
where $\bm{l}_j = \begin{bmatrix}
x_j &
y_j
\end{bmatrix}^\top$ is a vector of coordinates of $j$-th landmark.
The covariance matrices $\mathbf{R}_j$ are
given and typically determined by the precision of the sensor.
We conducted experiment with range measurements only (first output of $\bm{h}$),
bearing measurements only (second output of $\bm{h}$)
and both types of measurements.
% In some of the experiments we consider
% only range measurements
% (the first row in measurement model),
% in some of the experiments we have only
% bearing measurements (the second row in the measurement model)
% and in other experiments we have both
% range and bearing measurements.
The proposed approach is compared against model based
on linear time-variant stochastic differential equation (LTV SDE) 
\cite{barfoot}\footnote{The implementation was taken from 
\url{https://github.com/gtrll/gpslam}}.
LTV SDE is also based on GP with a special covariance matrix which has a band-diagonal inverse matrix, therefore,
its complexity
$\mathcal{O}(NM^2 + M^3)$.

The estimated trajectories are evaluated using
two metrics.
\begin{itemize}
    \item Absolute Pose Error (APE).
    This metric estimates global consistency of the trajectory.
    It is based on the relative pose on the estimated trajectory and ground truth trajectory:
    \[
        e_i^{abs} = \widehat{\mathbf{P}}_i \ominus \mathbf{P}_i =
        \left ( \mathbf{P}_i \right )^{-1} \widehat{\mathbf{P}}_i,
        \quad
        \mathbf{P}_i, \widehat{\mathbf{P}}_i \in SE(3),
    \]
    where $\mathbf{P}_i, \widehat{\mathbf{P}}_i$ are ground truth and estimated
    poses at time step $t_i$ represented by elements from $SE(3)$ group of rigid body transformations.
    We represent 2D points as 3D point by adding zero $z$-coordinate, roll and pitch angles.
    Then we compute the translational and rotational errors
    \[
    \begin{gathered}
        {\rm APE}_{trans} = \sqrt{\frac{1}{N} \|trans(e_i^{abs})\|_2^2},
        \\
        {\rm APE}_{rot} = \sqrt{\frac{1}{N} \|rot(e_i^{abs})\|_2^2},
    \end{gathered}
    \]
    where $trans(e)$ is a translational part of $e$ and
    $rot(e)$ is a rotational part of $e$.

    \item Relative Pose Error (RPE).
    This metric estimates the local consistency of the trajectory.
    It is invariant to drifts, i.e., if we translate and rotate the whole trajectory
    the RPE will remain the same.
    RPE is based on the relative difference of the poses on the estimated and
    ground truth trajectories:
    \begin{align*}
        e_i^{rel} = \hat{\delta}_i \ominus \delta_i =
        \left (
            \left ( \mathbf{P}_{i - 1} \right )^{-1} \mathbf{P}_i
        \right )^{-1}
        \left (
            \left ( \widehat{\mathbf{P}}_{i - 1} \right )^{-1} \widehat{\mathbf{P}}_i
        \right ),
    \end{align*}
    where $\mathbf{P}_i, \widehat{\mathbf{P}}_i \in SE(3)$ are as in APE.
    Similarly to APE, we calculate translational and rotational errors
    \[
    \begin{gathered}
        {\rm RPE}_{trans} = \sqrt{\frac{1}{N} \|trans(e_i^{rel})\|_2^2},
        \\
        {\rm RPE}_{rot} = \sqrt{\frac{1}{N} \|rot(e_i^{rel})\|_2^2}.
    \end{gathered}
    \]
\end{itemize}
We stress that our work is a proof of concept,
the approach was implemented purely in Python without any optimization.
Therefore, we do not provide any evaluation of the running time
of the algorithm.
We also note, that the computational complexity
is reasonably low (see Section \ref{sec:rff_slam}).
It depends on the number of features which sets a speed/accuracy trade-off.
Finding optimal number of features is a model selection
problem and it is out of the scope of the paper.

\subsubsection{Synthetic trajectories}
We generated $10$ different random trajectories, for each trajectory we conducted several experiments
with different noise level in observations,
different number of landmarks (from $5$ to $100$)
and different measurement types (range, bearing, range/bearing).
The noise was generated from Gaussian distribution with standard
deviation varying in $[1, 5]$ interval for range measurements
and bearing varying in $[1^\circ, 10^\circ]$ interval.

\paragraph*{Number of features $D$}
For the synthetic dataset we conducted experiments with different number of features $D$.
We observed that for a small $D$ ($D \sim 10$) the trajectory starts diverging when its length increases (at about $100$ observations).
Increasing the number of features increases the length of the trajectory for which the estimate does not diverge.
For the trajectories that we used in our experiments $D=100$ was
enough to obtain good state estimates.

\paragraph*{Kernel parameters}
The main kernel parameter is its lengthscale $\sigma_l$.
Typically, it affects the GP regression model the most.
It controls the smoothness of the obtained approximation.
Larger lengthscale should be used for smooth trajectories and smaller values for
less smooth trajectories.
In our experiments a rather wide kernel worked well,
we set $\sigma_l = 3.0$.
%An example of the estimated trajectories is given in Figure~\ref{fig:rand_2d}.
% In case of range only measurements there is no
% information about the heading of the robot
% in observation, so our approach cannot estimate it (it returns constant value).
% In this case we calculate the bearing as an angle of the movement direction.
The qualitative results can be found in Table~\ref{tab:rand_2d_errors}.
We can see that in the case of range and range-bearing measurements
the proposed approach looks more accurate.

\begin{figure}
    \centering
    \includegraphics[width=0.45\textwidth]{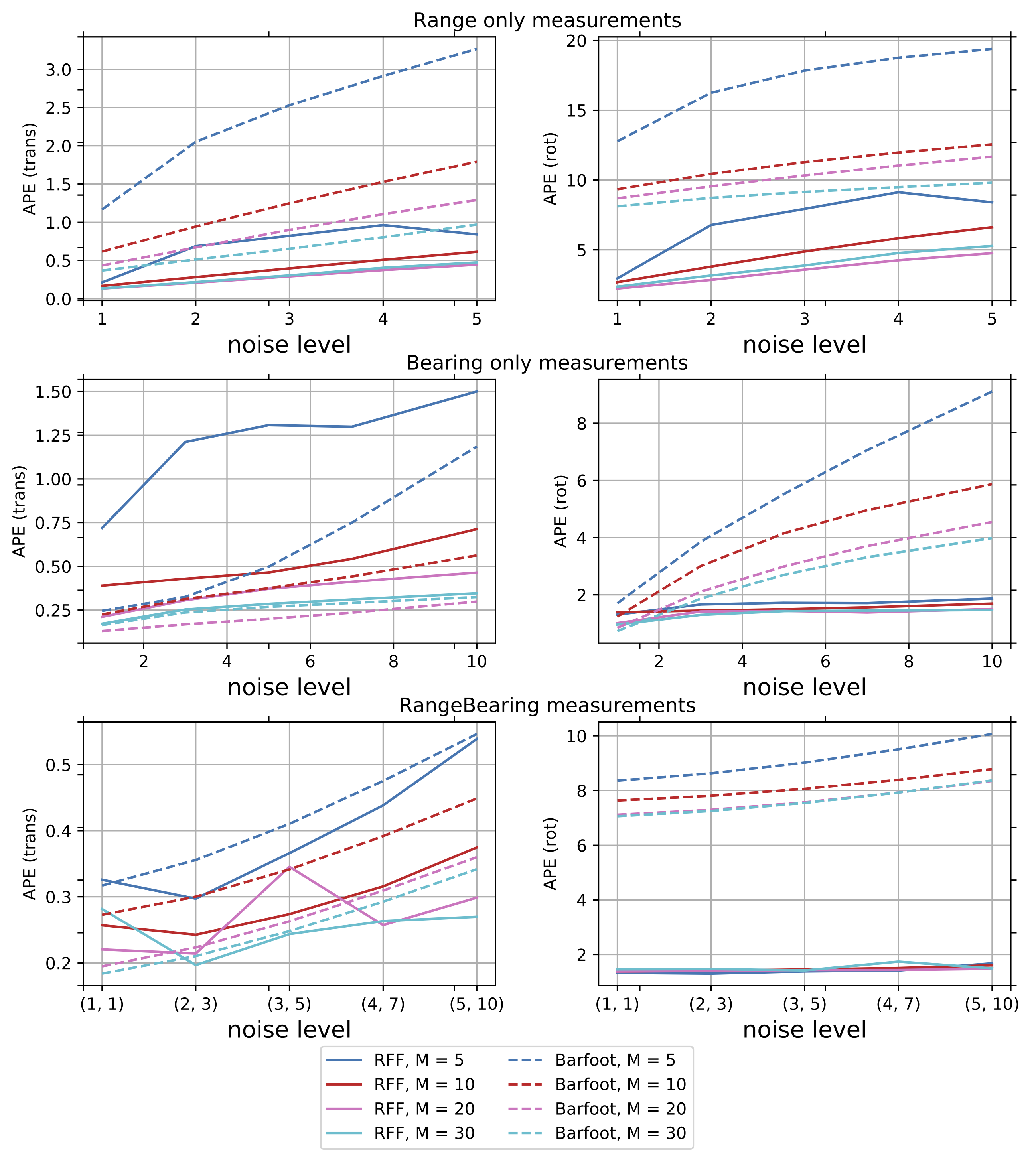}
    \caption{Average APE errors for synthetic trajectories at different noise levels and number of landmarks}
    \label{fig:synthetic_noise_distr}
\end{figure}

We also study the dependency of the estimation error on the noise level
and the number of landmarks.
In Figure~\ref{fig:synthetic_noise_distr}
you can see the APE translation errors
for different noise levels, the number of landmarks and measurement types.
We make several observations based on these results.
\begin{itemize}
    \item Our approach does not estimate bearing in the range-only measurements because there is no information about bearing in the data.
    In this case we calculate heading by
    calculating the movement direction of the estimated trajectory.
    Barfoot's method handles this situation due to their mean prior based on
    the differential equation.

    \item The proposed approach provides better rotation errors in all cases.

    \item The translation errors in range only
    setting and rotation errors of RFF 
    approach in bearing only measurements 
    increase slower with noise level compared 
    to LTV SDE errors.
    For range-bearing case the difference is less significant.
\end{itemize}
\begin{table}[]
    \centering
    \caption{Relative Errors for synthetic trajectories}
    \label{tab:rand_2d_errors}
    \begin{tabular}{cc|c c c}
     & & Pos. & Rot. & Landmarks \\
     \toprule
     RangeBearing & RFF & 0.022 & \textbf{0.154} & 6e-4 \\
     & LTV SDE & 0.025 & 5.602 & 0.110 \\
     \hline
     Range & RFF & \textbf{0.016} & \textbf{0.320} & 1e-6 \\
     & LTV SDE & 0.025 & 5.580 & 0.003 \\
     \hline
     Bearing & RFF & 0.035 & \textbf{0.152} & 8e-6 \\
     & LTV SDE & \textbf{0.025} & 1.200 & 0.016 \\
     \bottomrule
    \end{tabular}
\end{table}

\subsubsection{Autonomous Lawn-Mower}
\begin{figure}[h]
    \centering
    \includegraphics[width=0.45\textwidth]{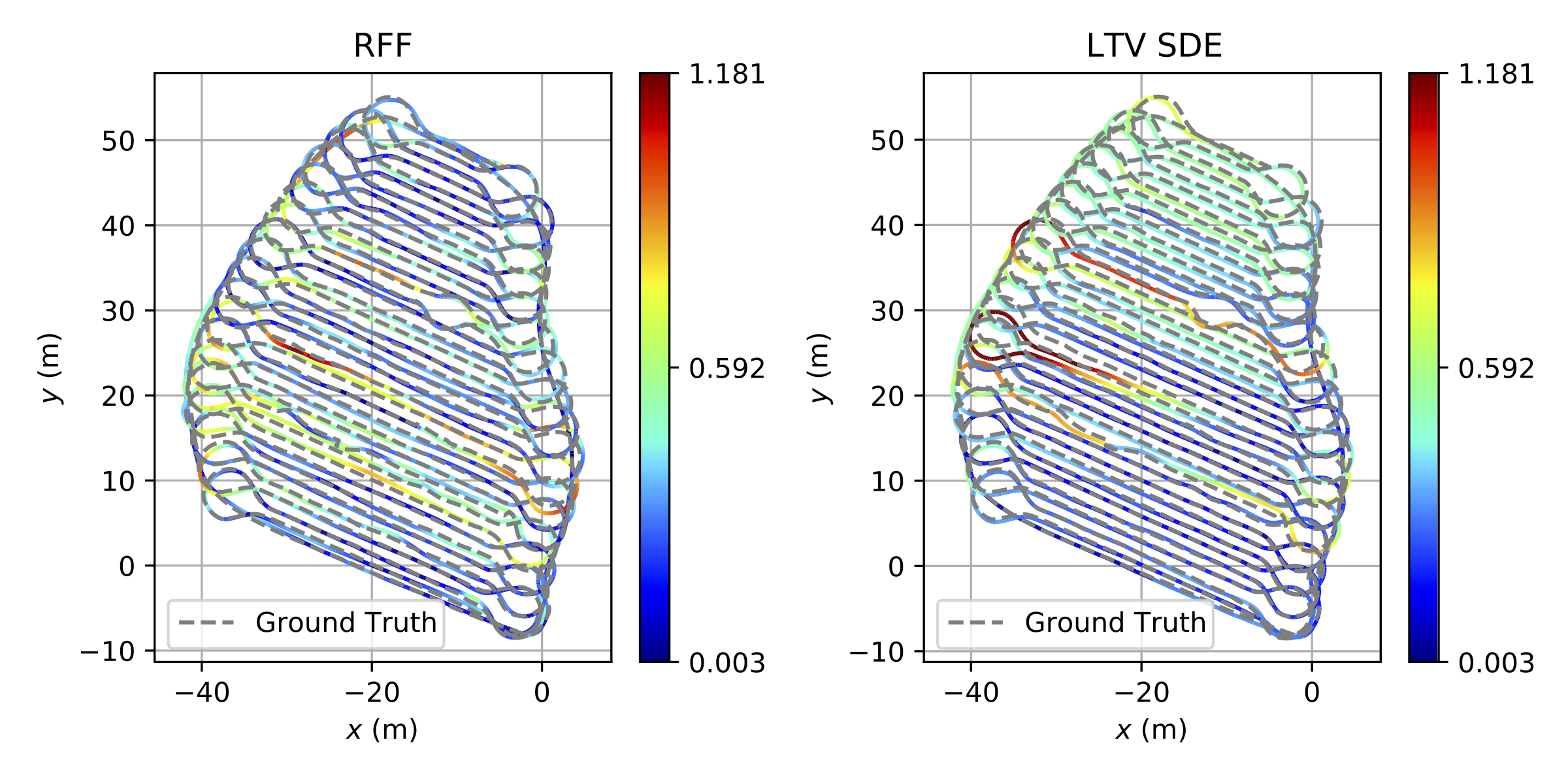}
    \caption{Distribution of APE errors along the trajecotry for Autonomous Lawn-Mower benchmark}
    \label{fig:autnomous_lawn_mower}
\end{figure}

In this experiment we evaluate our approach on a Plaza data set
collected from an autonomous
lawn-mower~\cite{djugash2010geolocation}.
The data set contains range measurements recorded using
time-of-flight and odometry measurements.
Odometry measurements come more frequently than range measurements.
The ground truth data was collected from GPS measurements
and according to \cite{djugash2010geolocation} its accuracy
is $2$cm.
The resulting trajectory is given in
Figure~\ref{fig:autnomous_lawn_mower}.
In this experiment we did batch updates with batch size $5$, i.e.
we updated the trajectory after each new $5$ measurements.
The motion model based prior was used as we have odometry
measurements.
We can see slight oscillations of the estimated trajectory.
They can be explained by the nature of the Fourier features.
However, the errors are comparable, see Table~\ref{tab:real_benchmarks_errors}.
The estimated trajectories are depicted in Figure~\ref{fig:autnomous_lawn_mower}.

% \subsubsection{Victoria Park dataset}
% We evaluate our approach on the Victoria Park dataset \cite{guivant2001optimization}.
% This dataset consists of range/bearing measurements and speed/steering odometry measurements of a vehicle that moves through
% the Victoria Park in Sydney.
% The trees in the park serve as landmarks.
% The total distance is $\sim3.5$ km, total time is about $26$ minutes.

% The data set contains odometry measurements, they are rather frequent, so we use motion model
% based prior for the mean trajectory.
% Note, that here we evaluate part of the trajectory for several reasons:
% first of all the data set is too large for online estimation (the complexity is quadratic in number of observations, the total number of observations including odometry is about $50000$).
% Secondly, our approach is written in pure Python and is not optimized for long trajectories.
% Also the benefits of the approach can be seen on short parts of the trajectory.
% The resulting trajectory is given in Figure~\ref{fig:victoria_park_rff}.

% \begin{figure}
%     \centering
%     \includegraphics[width=0.5\textwidth]{figures/victoria_park_rff.pdf}
%     \caption{Victoria Park dataset, estimated trajectory.}
%     \label{fig:victoria_park_rff}
% \end{figure}

\subsection{KITTI-projected dataset}

To evaluate our approach, we proceed with the famous dataset KITTI odometry\cite{kitti}. We used the dataset part with a sequence of stereo images taken while moving along a specific trajectory. The dataset also contains ground truth trajectory and camera information. 
To apply our approach, we extracted bearing observations by using visual SLAM from ORB-SLAM2 \cite{mur2017orb}.
From each stereo image we find keypoints with their coordinates in the local frame \cite{multiview}
which we use to calculate bearing observations. The keypoints in this case play the role of landmarks. The pipeline to project KITTI into a 2D dataset is following:

\begin{itemize}
\item Input: KITTI-odometry datasett(e.g. sequence 08);
\item Run ORB-SLAM2 to get observations (keypoints, timestamps);
\item Correct camera pose and keypoints by a transformation term that aligns with the vertical axis to make them independent of their original camera pose %\eqref{eq:corr}; 
\begin{equation}  \label{eq:corr}
R_t \cdot R_{correction} = \mathrm{Exp}([0,0,\alpha]^{\top});
\end{equation}
\item Calculate weights for each observations based on how much time this point was observed/visited;
\item Filter observations;
\item Do orthonormal projection for each observation;
\item Calculate bearing.
\end{itemize}

The number of landmarks (keypoints) is $80771$.
With such a big number of landmarks, the experiments are very slow, so we reduced the number of landmarks to $3975$.
We selected the landmarks randomly with probabilities proportional to their weights, but for each keyframe
we left not less than $10$ landmarks.
Also, to speed up the calculations we split the trajectory
into $10$ consecutive slices, do estimation on each slice
independently and then average the estimation errors.

The extracted data we then use in our approach to estimate the trajectory.
To check our assumption that kernels with dense precision 
matrices should work better in case of noisy observations,
we also generated a noisy version of the KITTI-projected
dataset.
To do so, we added Gaussian noise with standard deviation $\sigma$.

The results are given in Table~\ref{tab:real_benchmarks_errors}.
We can see that without additional noise the results are comparable (with RFF based approach being slightly better).
When we increase the amount of noise, absolute errors of our approach grow much slower compared to LTV SDE errors.

% \begin{table}[]
%     \footnotesize
%     \centering
%     \caption{Summary of the datasets}
%     \label{tab:datasets}
%     \begin{tabular}{c|c|c|c}
%          & \makecell{\# odometry\\ meas.} &
%          \makecell{\# range/bearing\\ meas.} & \# landmarks \\
%         \toprule
%         Synthetic & $0$ & $1000$ & $5-50$ \\
%         Lawn-Mower & $9658$ & $3529$ & $4$ \\
%         % Victoria Park & $44829$ & $7171$ & $637$ \\
%         KITTI & $0$ & $1379$ & $3975$ \\
%         \bottomrule
%     \end{tabular}
% \end{table}

\begin{table}[ht]
    \centering
    \caption{Real-world benchmark errors.}
    \label{tab:real_benchmarks_errors}
    \begin{tabular}{c|c|c|c|c}
    \multicolumn{5}{l}{}\\[-0.5em]
    \multicolumn{5}{c}{Autonomous Lawn-Mower} \\
    \multicolumn{5}{l}{}\\[-0.7em]
    \toprule
    & APE (trans) & APE (rot) & RPE (trans) & RPE (rot) \\
    \hline
    LTV SDE & $0.48$ & $1.44$ & $0.021$ & $0.10$ \\
    RFF & $0.42$ & $2.25$ & $0.026$ & $0.54$ \\
    \bottomrule
    
    \multicolumn{5}{l}{}\\[-0.5em]
    \multicolumn{5}{c}{KITTI-projected} \\
    \multicolumn{5}{l}{}\\[-0.7em]
    \toprule
    LTV SDE & $5.130$ & $1.059$ & $0.068$ & $0.113$ \\
    RFF & $5.070$ & $0.489$ & $0.040$ & $0.048$ \\
    \bottomrule

    \multicolumn{5}{l}{}\\[-0.5em]
    \multicolumn{5}{c}{KITTI-projected + noise, $\sigma=1^\circ$} \\
    \multicolumn{5}{l}{}\\[-0.7em]
    \toprule
    LTV SDE & $5.126$ & $1.086$ & $0.068$ & $0.133$ \\
    RFF & $5.070$ & $0.544$ & $0.0454$ & $0.052$\\
    \bottomrule

    \multicolumn{5}{l}{}\\[-0.5em]
    \multicolumn{5}{c}{KITTI-projected + noise, $\sigma=3^\circ$} \\
    \multicolumn{5}{l}{}\\[-0.7em]
    \toprule
    LTV SDE & $5.491$ & $3.136$ & $0.139$ & $0.259$ \\
    RFF & $5.075$ & $1.027$ & $0.073$ & $0.115$ \\
    \bottomrule
    
    \multicolumn{5}{l}{}\\[-0.5em]
    \multicolumn{5}{c}{KITTI-projected + noise, $\sigma=5^\circ$} \\
    \multicolumn{5}{l}{}\\[-0.7em]
    \toprule
    LTV SDE & $12.915$ & $5.114$ & $0.242$ & $0.358$\\
    RFF & $5.077$ & $1.304$ & $0.084$ & $0.119$ \\
    \bottomrule
    \end{tabular}
\end{table}

\section{CONCLUSION}
\label{sec:conclusion}
In this paper, we show how to apply GP for time-continuous SLAM with a less restricted class of kernel functions.
We accomplish it by using RFF approximation of the kernel.
The proposed approach has linear complexity in number of observations $N$, which is much faster compared to 
the traditional GP model.
The method provides a lot of flexibility for tuning various aspects
of the state estimate (smoothness, periodicity, etc.)
by choosing an appropriate kernel function.
However, such flexibility requires more careful tuning of the prior parameters (mainly, we need good prior mean for the trajectory
and careful balance between measurements covariance, weights and landmarks prior covariances).
We demonstrated the potential of the approach on synthetic and
real world datasets.
The experiments justify our assumption that in the case of 
noisy observations having dense inverse covariance matrix
helps to improve the accuracy compared to sparse matrices.

% \section*{Acknowledgment}
% Evgeny Burnaev was supported by the Russian Foundation for Basic Research grant 21-51-12005 NNIO\_a.

\addtolength{\textheight}{-10cm}   % This command serves to balance the column lengths
                                  % on the last page ofthe document manually. It shortens


% \section*{ACKNOWLEDGMENT}

\bibliographystyle{IEEEtran}
\bibliography{root}

\end{document}